\documentclass[conference]{IEEEtran}
\IEEEoverridecommandlockouts
\usepackage{cite}
\usepackage{amsmath,amssymb,amsfonts}
\usepackage{algorithmic}
\usepackage{graphicx}
\usepackage{textcomp}
\usepackage{xcolor}
\usepackage{lineno}
\usepackage{multirow}
\usepackage{longtable}
\usepackage{rotating}
\usepackage{color}
\usepackage{makecell}
\graphicspath{{figures/}}

\def\BibTeX{{\rm B\kern-.05em{\sc i\kern-.025em b}\kern-.08em
    T\kern-.1667em\lower.7ex\hbox{E}\kern-.125emX}}
\begin{document}

\title{Res2NetFuse: A Novel Res2Net-based Fusion Method for Infrared and Visible Images\\
\thanks{This work was supported by the National Natural Science Foundation of China (62202205, 62020106012), the Fundamental Research Funds for the Central Universities (JUSRP123030).}
}

\author{
\IEEEauthorblockN{1\textsuperscript{st} Xu Song}
\IEEEauthorblockA{\textit{School of Artificial Intelligence and Computer Science} \\
\textit{Jiangnan University}\\
Wuxi, China \\
xu\_song\_jnu@163.com}
\and

\IEEEauthorblockN{2\textsuperscript{nd} Yongbiao Xiao}
\IEEEauthorblockA{\textit{School of Artificial Intelligence and Computer Science} \\
\textit{Jiangnan University}\\
Wuxi, China \\
yongbiao\_xiao\_jnu@163.com}
\and

\IEEEauthorblockN{3\textsuperscript{rd} Hui Li*}
\IEEEauthorblockA{\textit{School of Artificial Intelligence and Computer Science} \\
\textit{Jiangnan University}\\
Wuxi, China \\
lihui.cv@jiangnan.edu.cn}
\and

\IEEEauthorblockN{4\textsuperscript{th} Xiao-Jun Wu}
\IEEEauthorblockA{\textit{School of Artificial Intelligence and Computer Science} \\
\textit{Jiangnan University}\\
Wuxi, China \\
xiaojun\_wu\_jnu@163.com}
\and

\IEEEauthorblockN{5\textsuperscript{th} Jun Sun}
\IEEEauthorblockA{\textit{School of Artificial Intelligence and Computer Science} \\
\textit{Jiangnan University}\\
Wuxi, China \\
sunjun\_wx@hotmail.com}
\and

\IEEEauthorblockN{6\textsuperscript{th} Vasile Palade}
\IEEEauthorblockA{\textit{ Faculty of Engineering, Environment and Computing } \\
\textit{Coventry University}\\
 Conventry, U.K. \\
ab5839@coventry.ac.uk}
}

\maketitle

\begin{abstract}
The fusion of visible light and infrared images has garnered significant attention in the field of imaging due to its pivotal role in various applications, including surveillance, remote sensing, and medical imaging. Therefore, this paper introduces a novel fusion framework using Res2Net architecture, capturing features across diverse receptive fields and scales for effective extraction of global and local features. Our methodology is structured into three fundamental components: the first part involves the Res2Net-based encoder, followed by the second part, which encompasses the fusion layer, and finally, the third part, which comprises the decoder. The encoder based on Res2Net is utilized for extracting multi-scale features from the input image. Simultaneously, with a single image as input, we introduce a pioneering training strategy tailored for a Res2Net-based encoder. We further enhance the fusion process with a novel strategy based on the attention model, ensuring precise reconstruction by the decoder for the fused image. Experimental results unequivocally showcase our method's unparalleled fusion performance, surpassing existing techniques, as evidenced by rigorous subjective and objective evaluations.
\end{abstract}


\section{Introduction}

The objective of image fusion is to combine information from several images to generate a composite image that retains the key features of each source image\cite{li2023lrrnet}. Image fusion finds applications in various fields, including remote sensing, medical imaging, and computer vision, where combining complementary information from different sensors or modalities enhances the overall understanding and analysis.
Specifically, visible and infrared image fusion is a challenging task \cite{10}. Merging the unique advantages of visible light and infrared imagery poses a notable challenge due to their inherent differences in wavelength and physical properties. Infrared images capture thermal information, enabling the visualization of heat signatures and subtle temperature variations, while visible light images provide rich color and detailed visual information. The amalgamation of these diverse datasets requires sophisticated techniques \cite{li2018infrared}\cite{li2020nestfuse} that preserve essential details from each modality, ensuring a unified, informative, and contextually relevant output.


Broadly, within the domain of image fusion, the multiscale transform (MST) method stands out as the prevailing approach. The fundamental stages of MST-based techniques encompass the decomposition stage, the fusion stage, and the reconstruction stage. Nevertheless, during the inverse transform process, MST-based methods tend to forfeit essential information inherent in the source images, consequently impacting the ultimate fusion outcomes\cite{3}.

In recent years, there have been significant advancements in image fusion methods driven by representation learning techniques. A comprehensive framework \cite{5} was proposed for image fusion, amalgamating SR and MST to address the inherent limitations of fusion methods based solely on SR or MST. Global and local saliency maps for fusion were derived from sparse coefficients of source images in sparse representation (SR), marking a significant approach \cite{4}. Another noteworthy approach, low-rank representation (LRR) was utilized in the image fusion \cite{6}, achieving superior performance in both local and global structures by concurrently integrating dictionary learning and LRR. Additionally, MDLatLRR \cite{li2020mdlatlrr}, a new decomposition framework with latent LRR, was applied for image fusion.

For deep learning, a plethora of image fusion techniques grounded in deep learning principles have emerged. 
DenseFuse\cite{10}, an innovative image fusion framework, was introduced with a strong emphasis on extensively utilizing middle-layer features. 
Its architecture comprises an encoder, a fusion layer, and a decoder.
Advancing this approach, MSDNet \cite{11} was proposed for medical image fusion. The authors augmented Li's \cite{10} encoder with a multi-scale layer, resulting in a highly effective fusion process. Nowadays, in the realm of image fusion, CUFD \cite{xu2022cufd} employs dual sets of encoder-decoder networks. Additionally, a novel image fusion network with an end-to-end architecture has been introduced, denoted as MUFusion \cite{cheng2023mufusion}.

However, existing algorithms tend to overlook the constraints posed by the receptive fields of the convolutional neural network, resulting in deficiencies in the extraction of the features at different scales. 
In order to address this problem, this paper incorporates the Res2Net concept \cite{12} into the encoder. By utilizing this way, we can extract multi-scale features. 
Section \ref{Res2Net} will provide a detailed discussion. 
As follows, the primary contributions of this paper are outlined:

(1) Res2Net-based Encoder: Our method utilizes the Res2Net module in the encoder, enabling the extraction of multi-scale features from source images.

(2) Novel Training Strategy: A simple yet efficient training strategy is introduced to train our framework, achieving comparable performance to models trained on larger datasets, and significantly reducing training time.

(3) Enhanced Fusion Strategy: We introduce an improved fusion strategy leveraging attention mechanisms, ensuring more precise feature fusion.

(4) Comprehensive Evaluation: We conduct thorough subjective and objective evaluations, demonstrating the superior performance of our approach compared to existing methods.

Following sections of this paper are structured as: the concise overviews of DenseFuse and Res2Net are depicted in Section \ref{relatedworks}. The novel fusion method is proposed in Section \ref{Methodology}. The Experimental results are shown in Section \ref{experiment}. The conclusion is in Section \ref{conclusion}.

\section{Related Works}
\label{relatedworks}

\subsection{DenseFuse}
\label{DenseFuse}

\begin{figure}[!ht]
\centering
\includegraphics[width=0.8\linewidth]{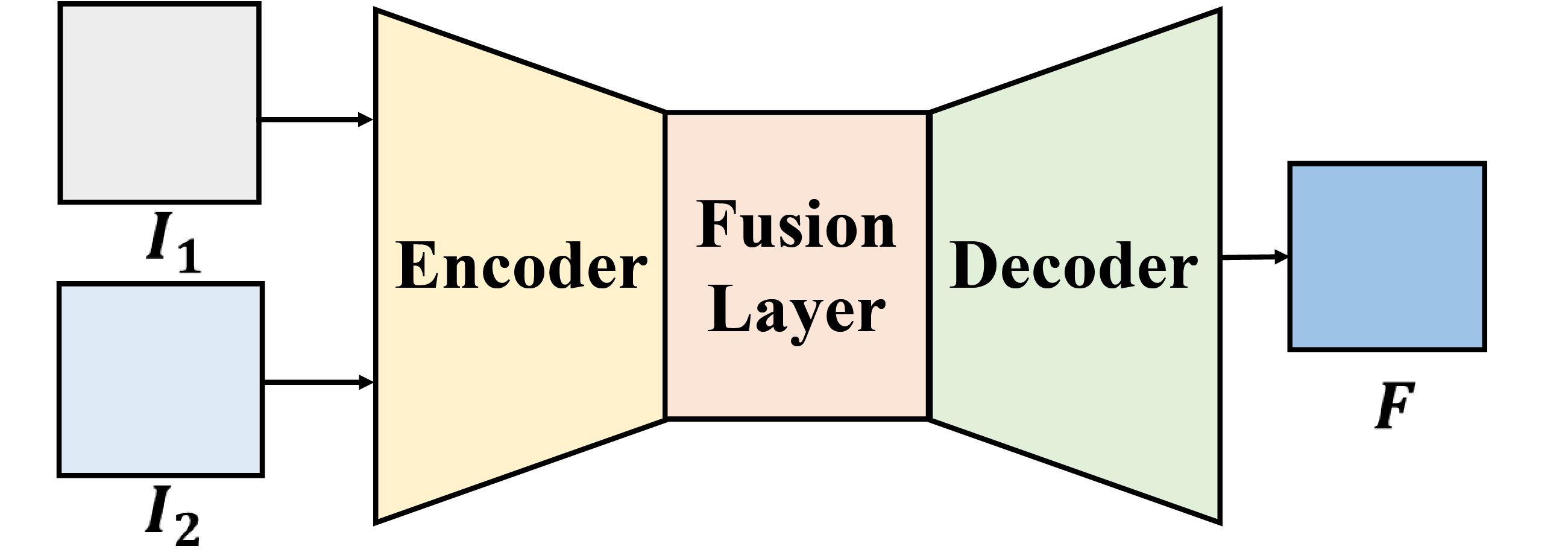}
\caption{DenseFuse Architecture.}
\label{densefuse}
\end{figure}

In the DenseFuse architecture, as outlined in \cite{10}, there are three core components: the encoder component, the fusion component, and the decoder component. During the training phase, both the encoder and decoder are employed to reconstruct input images. This approach enables the training dataset to cover diverse image types, enhancing the model's adaptability. Subsequent to training, the encoder gains feature extraction abilities, while the decoder specializes in image reconstruction. Visualize the DenseFuse architecture in Fig.~\ref{densefuse} for a detailed overview.

As Fig.~\ref{densefuse} shows, the fusion layer combines two feature sets extracted during the encoding phase, facilitating the subsequent reconstruction of the fused image by the decoder.

\subsection{Res2Net}
\label{Res2Net}

\begin{figure}[!ht]
\centering
\includegraphics[width=0.8\linewidth]{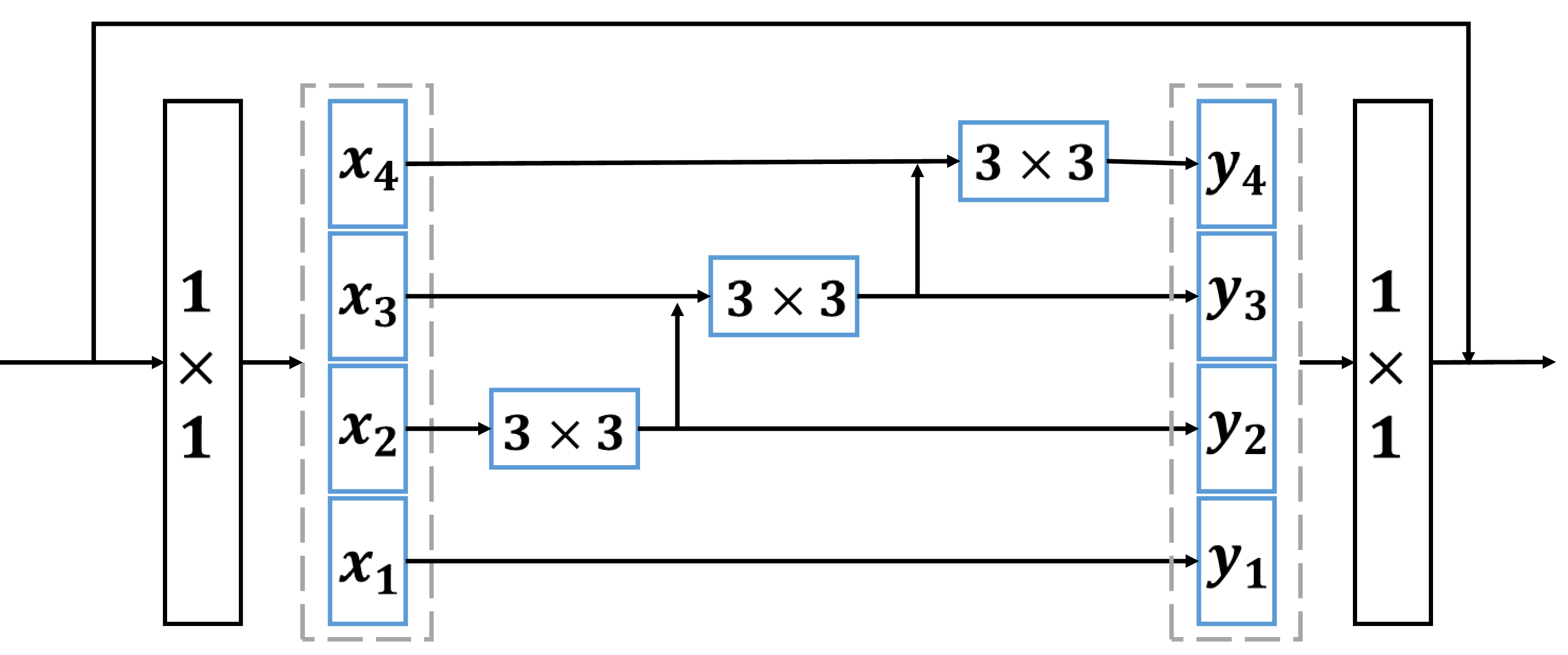}
\caption{Res2Net module.}
\label{Res2Net module}
\end{figure}

Multi-scale features are pivotal in various image processing tasks. Gao et al. \cite{12} introduced Res2Net, a novel building block for convolutional neural networks, as depicted in Fig.~\ref{Res2Net module}.

As shown in Fig.~\ref{Res2Net module}, post the $1\times1$ filtering, the feature maps undergo an even division, forming subsets denoted as $x_i$, where $i\in{1,2, \ldots,n}$ with n representing the count of subsets. Each subset, excluding $x_1$, undergoes a specific filtering operation symbolized as $F_i()$. The resulting convolution output is termed as $y_i$, and its computation is elaborated in \eqref{equ:$y_i$}.
\begin{align}\label{equ:$y_i$}
  	y_i=\left\{\begin{array}{ll}
		x_i & \textrm{ $i=1$ }  \\
          F_i(x_i) & \textrm{ $i=2$  } \\
		F_i(x_i+y_{i-1}) & \textrm{$2 <i\leq n$}
	\end{array}\right.
\end{align}

From equation \eqref{equ:$y_i$}, we can see that when $x_i$ undergoes a $3\times3$ filtering process, the output ($y_i$) attains a broader receptive field in contrast to $x_i$. The next phase involves implementing the fundamental concept of Res2Net, which involves creating hierarchical residual-like connections within a residual block. This strategy adeptly captures multi-scale features across diverse granularities, thus enhancing the receptive field across network layers.

\section{Methodology}
\label{Methodology}

In this section, we provide a comprehensive explanation of the proposed fusion method for grayscale images. This section is organized into three subsections: Section \ref{proposedmethod} delves into the intricacies of the proposed method, Section \ref{Training} elucidates the training network, encompassing aspects such as the loss function and training strategy, and Section \ref{fusionstrategy} expounds on the fusion layer, outlining the adopted strategy.

\begin{figure*}[!ht]
\centering
\includegraphics[width=0.75\linewidth]{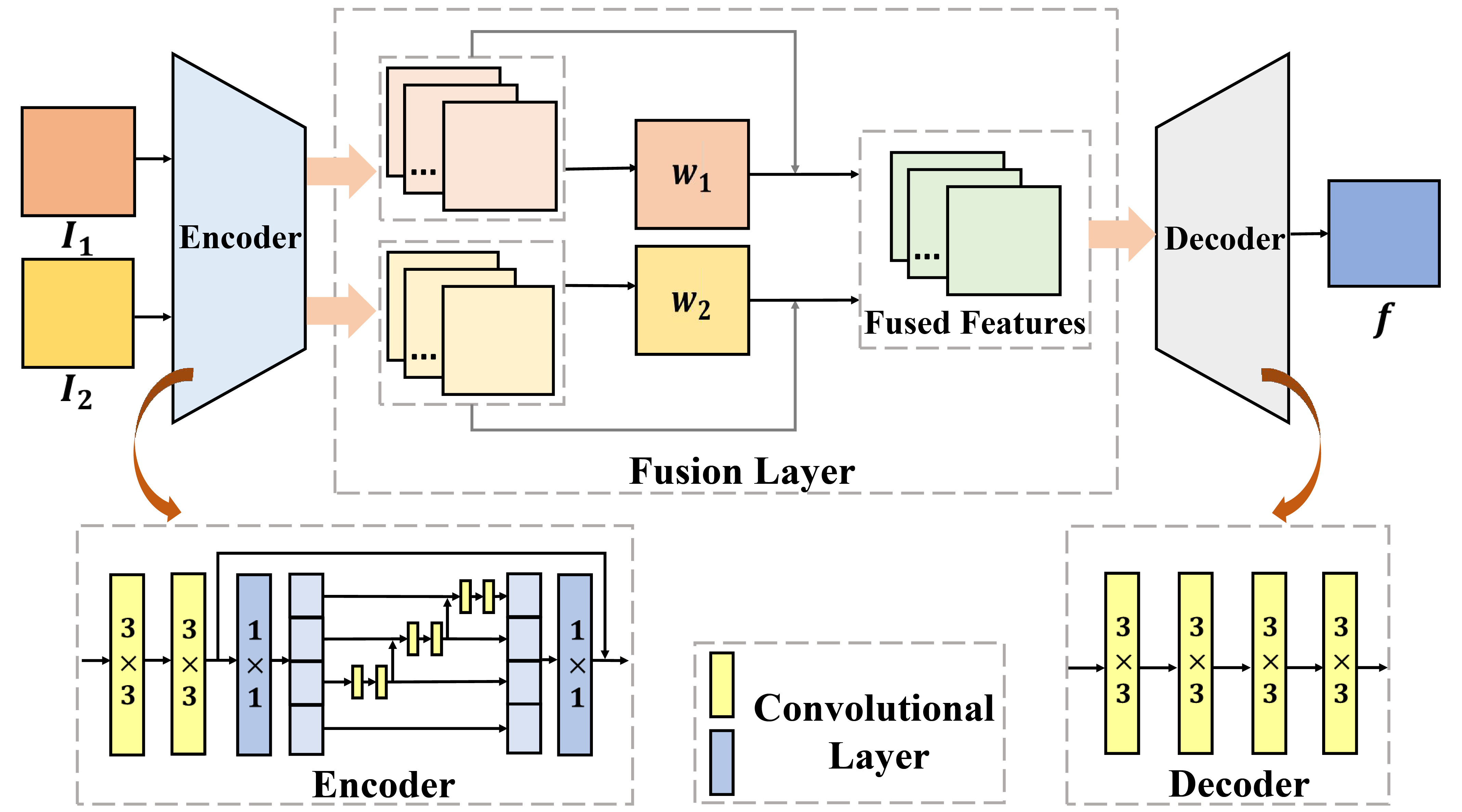}
\caption{The presented framework is tailored for grayscale images. $I_1$ and $I_2$ mean the source image, both of which undergo encoding to extract features at different scales. Subsequently, in the fusion layer, these extracted features are fused before being fed into the decoder. Ultimately, the output of the decoder is the fused image denoted as $f$.}
\label{framework}
\end{figure*}


\subsection{The Proposed Fusion Method} 
\label{proposedmethod}

Motivated by the remarkable achievements of Res2Net and DenseFuse architectures, the Res2Net architecture is applied in our encoder due to its exceptional capability in extracting multi-scale features.

The input images, denoted as $I_1$ and $I_2$, represent infrared and visible images. Our proposed architecture comprises three core modules: the encoder module, the fusion module, and the decoder module. In Fig.~\ref{framework}, the schematic representation of our method can be observed. 

The encoder incorporates two $3\times3$ convolutional layers and a Res2Net block. At the start of the Res2Net block, a $1\times1$ convolution is applied, evenly dividing all features. These divisions are then passed through hierarchical residual-like connections and 3x3 convolutional layers. Subsequently, the processed divisions are concatenated. Ultimately, at the conclusion of the Res2Net block, the features are extracted using a $1\times1$ filter. The specific equation and detailed operation for the Res2Net block are elaborated in Section \ref{Res2Net}. Meanwhile, the decoder is composed of four $3\times3$ convolutional layers. In addition, the fusion strategy will be explained later, which is utilized in the fusion layer.

\subsection{Training}
\label{Training}

Whether grayscale or color images are employed, training network architecture remains consistent, comprising two fundamental components: the encoder and decoder. Known as a reconstruction model, primary function of the training network is to recreate the input image. The training network (including encoder \& decoder) parameters are detailed in Table \ref{tab:outline}.

\begin{table}[!ht]
\centering
\caption{\label{tab:outline}Training network structure. \textbf{Conv1} and \textbf{Conv3} represent $1\times1$ and $3\times3$ convolutional layers, respectively. Additionally, \textbf{(Conv3)$\times2$} indicates two identical layers.}
\resizebox{3.5in}{!}{
\begin{tabular}{|c|c|c|c|c|c|c|}
\hline
 & Layer & Size & Stride & \makecell[cc]{Channel\\(Input)} & \makecell[cc]{Channel\\(Output)} & Activation \\
\hline
\multirow{3}*{Encoder} &
Conv3      & 3      & 1      & 1      & 32      & ReLU \\
\cline{2-7}
~ & Conv3  & 3      & 1     & 32      & 64      & ReLU\\
\cline{2-7}
~ & Res2Net Block   &  --   &    --   &    --  &  --    & --\\
\cline{2-7}
\hline
\multirow{4}*{Decoder} &
    Conv3     & 3      & 1      & 64     & 64      & ReLU \\
\cline{2-7}
~ & Conv3     & 3      & 1      & 64     & 32      & ReLU \\
\cline{2-7}
~ & Conv3     & 3      & 1      & 32     & 16      & ReLU \\
\cline{2-7}
~ & Conv3     & 3      & 1      & 16     & 1       & --\\
\cline{2-7}
\hline
\multirow{6}*{\makecell[cc]{Res2Net Block}} &
    Conv1     & 1      & 1      & 64     & 64      & ReLU \\
\cline{2-7}
~ &     --      &   --     &    --   & --     & 16      &   --   \\
\cline{2-7}
~ & (Conv3)$\times2$   & 3      & 1      & 16     & 16      & ReLU\\
\cline{2-7}
~ & (Conv3)$\times2$   & 3      & 1      & 16     & 16      &ReLU\\
\cline{2-7}
~ & (Conv3)$\times2$   & 3      & 1      & 16     & 16      &ReLU\\
\cline{2-7}
~ & Conv1     & 1      & 1      & 64     & 64      & ReLU \\
\hline
\end{tabular}}
\end{table}

\textbf{Reconstruction Loss}. In order to achieve the reconstruction of the input image, the loss function \cite{10} we employ is depicted in \eqref{equ:lossfunction}.

\begin{equation} \label{equ:lossfunction}
L = L_{ssim} + L_{pixel}
\end{equation}

The $L_{pixel}$ and the $L_{ssim}$ are respectively calculated as:
\begin{equation}\label{L_ssim}
	L_{ssim} = 1 - SSIM(O,I)
\end{equation}
\begin{equation}\label{L_pixel}
	L_{pixel} = \frac{1}{BCHW}||O-I||^2_2
\end{equation}
where $B$ represents the batch size, $I$ and $O$ represent the input and output images, $C$ signifies the number of channels of $O$, $W$ and $H$ represent the width and height of $O$. Furthermore, SSIM($\cdot$) represents the structural similarity index \cite{14}.

\textbf{Training Strategy}. In the study by \cite{15}, it was established that the internal statistics of patches within a natural image offer sufficient information to train a robust generative model.
In the study by \cite{15}, it demonstrated that the statistical characteristics of patches within a natural image contain enough information to develop a resilient generative model.
Therefore, our approach involves training a generative model (auto-encoder) using a single natural image at various scales, avoiding the reliance on multiple samples from a database.

Initially following the methodology outlined in \cite{15}, we trained the reconstruction model by a single natural image at different scales. However, experimental results revealed a slight blurring of detailed information in the output image, despite the model's ability to reconstruct the input image. To address this issue, we trained the reconstruction model 2000 times using an entire natural image. Surprisingly, the results showed that our model's reconstruction proficiency was comparable to that of DenseFuse \cite{10} and MSDNet \cite{11}, both trained with 80,000 images, while significantly reducing the training time. In Section \ref{Reconstruction}, a detailed evaluation and analysis of this training strategy will be presented.


\subsection{The Fusion Layer}
\label{fusionstrategy}

Following the training phase, the fusion process takes place within the intermediary layer nestled between the encoding network (the encoder) and the decoding network (the decoder). Within this fusion layer, the blending of features plays a crucial role in image fusion, requiring an intricate strategy.

Given the inherent disparities in the features extracted from input images, two distinct feature sets emerge after the encoding stage. To effectively merge these features, the conventional average strategy proves insufficient due to its simplistic nature. Therefore, as depicted in Fig.~\ref{framework}, our proposed approach integrates a spatial attention model into the fusion layer, providing a more sophisticated and nuanced method for feature integration.

\textbf{(1) Spatial Attention Utilizing $l_1$-norm. } Drawing inspiration from this\cite{10}, $l_1$-norm was employed to handle the features extracted by the encoder. Following this processing step, two essential weight maps represented as $w_1$ and $w_2$, are computed through the formula outlined in \eqref{equ:weightmap1}.
\begin{equation}\label{equ:weightmap1}
  	w_i(x,y) = \frac{L_1(\phi_i^{1:m}(x,y))}{\sum_{i=1}^kL_1(\phi_i^{1:m}(x,y))}
\end{equation}
where $L_1(\cdot)$ signifies the $l_1$-norm, and $\phi_i^{1:m}$ refers to the features extracted by the encoder, where $k$ denotes the number of input images, and $m$ signifies the number of feature maps. Specifically, we set the value of $k$ to 2.

In the final step, the improved fused features, denoted as $f^{1:m}$, are calculated utilizing the equation provided in \eqref{equ:fusedmaps1}.
\begin{equation}\label{equ:fusedmaps1}
  	f^{1:m}=\sum_{i=1}^kw_i\times\phi_i^{1:m}
\end{equation}

\textbf{(2) Spatial Attention Utilizing Mean Operation.}    
To initiate the process, the features $\phi_i^{1:m}$ extracted by the encoder undergo a mean operation. Subsequently, the softmax operation is employed, yielding the weight maps $w_1$ and $w_2$ as per the formula in \eqref{equ:weightmap2}.
\begin{equation}\label{equ:weightmap2}
  	w_i(x,y) = \frac{M(\phi_i^{1:m}(x,y))}{\sum_{i=1}^kM(\phi_i^{1:m}(x,y))}
\end{equation}
where mean operation, represented as $M(\cdot)$, is applied to individual pixels at coordinates $(x,y)$ within each feature map. Subsequently, according to the formula \eqref{equ:fusedmaps2}, the fused features $f^{1:m}$ are computed.
\begin{equation}\label{equ:fusedmaps2}
  	f^{1:m}=\sum_{i=1}^kw_i\times\phi_i^{1:m}
\end{equation}

Upon traversing the fusion layer, we obtain the ultimate fused features denoted as $f^{1:m}$, which are subsequently forwarded to the decoder. Ultimately, the final fused image is reconstructed using these amalgamated features.

\section{Experiments}
\label{experiment}
We begin with an overview of our experimental setup and environment. We then delve into a detailed performance analysis of our single-image training network. Following that, we present the experimental results.

\subsection{Environment and Experimental Settings}
\label{settings}

We employed 20 pairs of test images, comprising both visible and infrared images. We compared our method with several others, such as FusionGAN\cite{22}, DenseFuse\cite{10}, JSRSD \cite{4}, JSR\cite{20}, DCHWT\cite{19}, AEFusion\cite{li2023aefusion}, AT-GAN\cite{rao2023gan}, CUFD\cite{xu2022cufd}, MUFusion\cite{cheng2023mufusion}, and MSDNet\cite{11}. To objectively assess the fusion results, we considered eight indices:
\begin{itemize}
\item SCD\cite{28}: Measures transferred information from the input image to the final fused image.
\item MS\_SSIM\cite{29}: Serves as a no-reference quality assessment for image fusion.
\item $Q_{abf}$\cite{25}: Assesses the quality of visual information.
\item $FMI_w$ and $FMI_{dct}$\cite{26}: Calculate feature mutual information, such as wavelet and discrete cosine features.
\item $SSIM_a$\cite{14}: Calculated using \eqref{ssim}, measuring the structural information of the source images. In \eqref{ssim}, SSIM($\cdot$) means the structural similarity operation and $f$ represents the final fused image.
\end{itemize}
\begin{equation}\label{ssim}
SSIM_a(f) = (SSIM(f,I_1)+SSIM(f,I_2))\times 0.5
\end{equation}

In the training phase, we conducted 2000 iterations with a batch size of 1 and employed a single grayscale training image sized at $256 \times 256$. Our method is implemented using PyTorch, a powerful deep learning framework, and accelerated with an NVIDIA GTX 1050Ti GPU for optimized performance.

\subsection{Training Strategy}
\label{Reconstruction}
Natural images inherently display robust predictive power\cite{glasner2009super} and significant internal data redundancy\cite{shocher2018zero}. Inspired by these findings, we harnessed the combined potential of internal information and deep learning's generalization capabilities. We trained our reconstruction model using a single natural image, repeating this process 2000 times.
\begin{table*}[!ht]
\centering
\caption{\label{tab:fusionl1mean}Average evaluation metric results of 20 fused images. Specifically, \textbf{\emph{$l_1$-norm}} and \textbf{\emph{mean}} representing the fusion strategy. Color-coded for clarity: the best results utilize \textcolor{red}{Red}, the second best results utilize \textcolor{blue}{blue}, and the third best results utilize \textcolor{green}{green}}
\resizebox{0.7\textwidth}{!}{
\begin{tabular}{|c|c|c|c|c|c|c|c|}
\hline
\multicolumn{2}{|c|}{Methods} & $SSIM_a$  & $Q_{abf}$ & $FMI_{dct}$ & $FMI_w$ & SCD & MS\_SSIM \\
\hline
\multicolumn{2}{|c|}{DenseFuse} & 0.72829  & 0.43454 & {\color{red}{\textbf{0.41456}}} & {\color{green}{\textbf{0.42525}}} & {\color{blue}{\textbf{1.83379}}} & {\color{red}{\textbf{0.92860}}}\\
\hline
\multirow{2}*{MSDNet}&
	$l_1$-norm & {\color{blue}{\textbf{0.76365}}} & 0.44416 & 0.37547 & 0.41741 & 1.61825 & 0.83311 \\
\cline{2-8}
~& Mean & {\color{red}{\textbf{0.77453}}}  & 0.39214 & 0.39046 & 0.41116 & 1.67953 & 0.87041 \\
\cline{2-8}
\hline
\multirow{2}*{Res2NetFuse\_80000}&
	$l_1$-norm & 0.70881  & {\color{blue}{\textbf{0.47354}}} & 0.37620 & {\color{red}{\textbf{0.43160}}} & 1.67887 & 0.84166\\
\cline{2-8}
~& Mean & 0.72993  & 0.45175 & {\color{blue}{\textbf{0.40685}}} & 0.42417 & {\color{green}{\textbf{1.82024}}} & {\color{green}{\textbf{0.91695}}}\\
\hline
\multirow{2}*{Res2NetFuse\_1}&
	$l_1$-norm & 0.72520  & {\color{red}{\textbf{0.48364}}} & 0.37552 & {\color{blue}{\textbf{0.42924}}} & 1.69888 & 0.85384 \\
\cline{2-8}
~& Mean & {\color{green}{\textbf{0.74206}}}  & {\color{green}{\textbf{0.46368}}} & {\color{green}{\textbf{0.40262}}} & 0.42334 & {\color{red}{\textbf{1.83464}}} & {\color{blue}{\textbf{0.92129}}} \\
\hline
\end{tabular}}
\end{table*}

\begin{figure}[!ht]
\centering
\includegraphics[width=\linewidth]{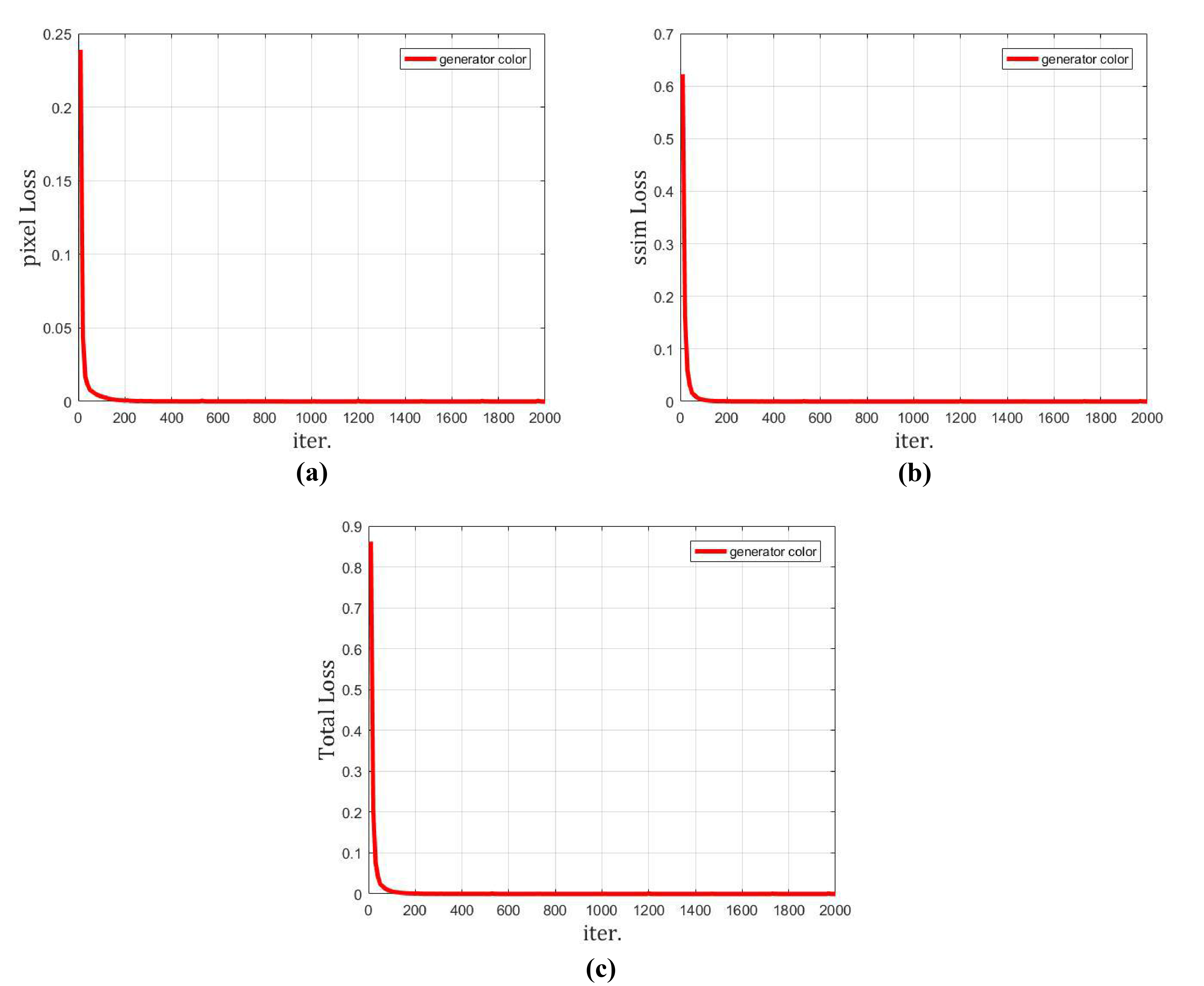}
\caption{Loss graphs. (a) $L_{pixel}$; (b) $L_{ssim}$; (c) Total loss.}
\label{loss}
\end{figure}

To illustrate our training progress, we present the loss graph in Fig.~\ref{loss}, which shows that the training model stabilizes after 200 iterations. Subsequently, we provide comparison results to demonstrate the trained model's effectiveness in terms of both reconstruction and fusion.

\begin{table}[!ht]
\centering
\caption{\label{tab:reconstruction} Average evaluation metric results for 20 reconstructed images, including \textbf{\emph{DenseFuseRecon}} (DenseFuse's reconstruction model), \textbf{\emph{MSDNetRecon}} (MSDNet's reconstruction model), \textbf{\emph{Res2NetFuseRecon\_1}} (Res2NetFuse's reconstruction model trained by one image), and \textbf{\emph{Res2NetFuseRecon\_80000}} (Res2NetFuse's reconstruction model trained by 80000 images)}
\resizebox{3.5in}{!}{
\begin{tabular}{|c|c|c|c|}
\hline 
Methods & MSE & SSIM & PSNR  \\
\hline
DenseFuseRecon & 0.00016 & 0.99560 & 43.62612 \\
\hline
MSDNetRecon & {\color{red}{\textbf{0.00001}}} & {\color{red}{\textbf{0.99933}}} & {\color{red}{\textbf{52.10881}}} \\
\hline  
Res2NetFuseRecon\_1 & 0.00042 & 0.99560 & 40.36431 \\
\hline
Res2NetFuseRecon\_80000 & 0.00005 & 0.99888 & 46.50366 \\
\hline
\end{tabular}}
\end{table}

\subsubsection{Reconstruction Comparisons} To evaluate the performance of our reconstruction model, we compared it with existing models trained on 80,000 images, including DenseFuse, MSDNet, and our proposed method. We used 20 images from MS-COCO for reconstruction experiments, utilizing peak signal-to-noise ratio (PSNR), structural similarity (SSIM), and mean squared error (MSE) (computed as shown in equation \eqref{MSE}) as assessment criteria.
\begin{equation}\label{MSE}
	MSE = \frac{1}{W\times H}\sum_{x=1}^W\sum_{y=1}^H(I(x,y)-R(x,y))^2
\end{equation} 
where $H$ and $W$ represent the image height and the image width, respectively, $I$ denotes the input image, and $R$ represents the reconstructed image.

Table \ref{tab:reconstruction} displays thes comparison results. As we can see, Res2NetFuse still exhibits comparable reconstruction capability, although it may not achieve the highest metrics. In contrast to Res2NetFuseRecon\_80000, the model trained on a single image attains both acceptable and comparable reconstruction performance, emphasizing the sufficiency of internal information from a single image for our lightweight network. Additionally, this training strategy significantly reduces the training time.

\begin{figure*}
\centering
\includegraphics[width=0.9\linewidth]{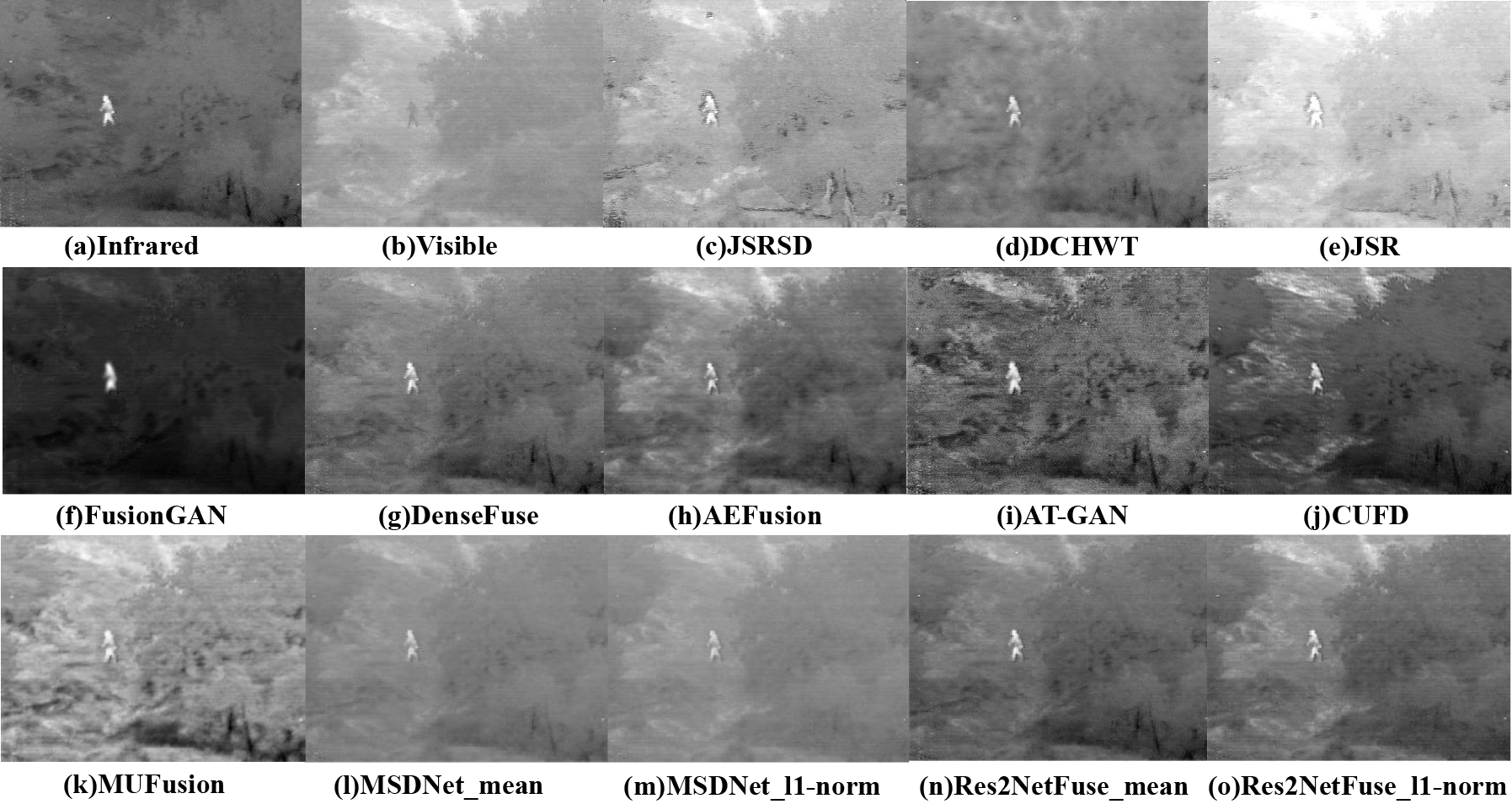}
\caption{Experimental findings: (a) The infrared image (b) The visible light image (c) JSRSD (d) DCHWT (e) JSR (f) FusionGAN (g) DenseFuse (h) AEFusion (i) AT-GAN (j) CUED (k) MUFusion (l) MSDNet utilizing the mean fusion strategy (m) MSDNet utilizing the $l_1$-norm fusion strategy (n) Res2NetFuse utilizing the mean fusion strategy (o) Res2NetFuse utilizing the $l_1$-norm fusion strategy.}
\label{results2}
\end{figure*}
\begin{table*}[!htpb]
\centering
\caption{\label{tab:ob}The 20 fused images are evaluated based on average values, color-coded for clarity: The best results utilize \textcolor{red}{Red}, the second best results utilize \textcolor{blue}{blue}, and the third best results utilize \textcolor{green}{green}.}
\resizebox{0.7\textwidth}{!}{
\begin{tabular}{|c|c|c|c|c|c|c|c|}
\hline
\multicolumn{2}{|c|}{Methods} & $SSIM_a$  & $Q_{abf}$ & $FMI_{dct}$ & $FMI_w$ & SCD & MS\_SSIM \\
\hline
\multicolumn{2}{|c|}{DCHWT} & 0.73078   & {\color{green}{\textbf{0.45890}}} & 0.38061 & 0.39700 & 1.61007 & 0.84278 \\
\hline
\multicolumn{2}{|c|}{JSR} & 0.60912  & 0.36267  & 0.16738 & 0.21284 & {\color{green}{\textbf{1.75518}}} & 0.84735 \\
\hline
\multicolumn{2}{|c|}{JSRSD} & 0.54471  & 0.32914  & 0.14560 & 0.18697 & 1.59142 & 0.76548 \\
\hline
\multicolumn{2}{|c|}{DenseFuse} & 0.72829  & 0.43454 & {\color{red}{\textbf{0.41456}}} & {\color{blue}{\textbf{0.42525}}} & {\color{blue}{\textbf{1.83379}}} & {\color{red}{\textbf{0.92860}}}\\
\hline
\multicolumn{2}{|c|}{FusionGAN} & 0.65207  & 0.21853 & 0.36097 & 0.36797 & 1.45818 & 0.73233 \\
\hline
\multicolumn{2}{|c|}{AEFusion} & 0.68948  &0.32510 &0.17045 &0.22482 &1.75508 &0.86873 \\
\hline
   
\multicolumn{2}{|c|}{AT-GAN} & 0.56636  &0.35253 &0.36184 &0.39656 &1.68997 &0.86337\\
\hline

\multicolumn{2}{|c|}{CUFD} & 0.64916  &0.37775 &0.19354 &0.26088 &1.61723 &0.78413 \\
\hline

\multicolumn{2}{|c|}{MUFusion} & 0.65392  &0.37838 &0.23752 &0.28688 &1.6934 &0.86064 \\
\hline

\multirow{2}*{MSDNet}&
	$l_1$-norm & {\color{blue}{\textbf{0.76365}}} & 0.44416 & 0.37547 & 0.41741 & 1.61825 & 0.83311 \\
\cline{2-8}
~& Mean & {\color{red}{\textbf{0.77453}}}  & 0.39214 & {\color{green}{\textbf{0.39046}}} & 0.41116 & 1.67953 & {\color{green}{\textbf{0.87041}}} \\
\cline{2-8}
\hline
\multirow{2}*{Res2NetFuse}&
	$l_1$-norm & 0.72520  & {\color{red}{\textbf{0.48364}}} & 0.37552 & {\color{red}{\textbf{0.42924}}} & 1.69888 & 0.85384 \\
\cline{2-8}
~& Mean & {\color{green}{\textbf{0.74206}}}  & {\color{blue}{\textbf{0.46368}}} & {\color{blue}{\textbf{0.40262}}} & {\color{green}{\textbf{0.42334}}} &{\color{red}{\textbf{1.83464}}} & {\color{blue}{\textbf{0.92129}}} \\
\hline
\end{tabular}}
\end{table*}

\subsubsection{Fusion Comparisons} In this section, we assess Res2NetFuse\_1, trained on a single image, alongside DenseFuse, MSDNet, and Res2NetFuse\_80,000, trained with 80,000 images. MSDNet also adopts a multi-scale approach, ensuring consistent fusion strategies during comparison, in Table \ref{tab:fusionl1mean} (the color-coding indicates performance rankings).
 
Res2NetFuse\_1 exhibits comparable fusion capabilities to the other methods. Notably, while MSDNet outperforms Res2NetFuse in reconstruction, Res2NetFuse achieves superior fusion results under the same strategy. This difference can be attributed to MSDNet's activity level maps being specific to different scales, whereas Res2NetFuse's maps cover all scales, resulting in more stable weight maps.

Comparing with Res2NetFuse\_80,000, we observe similar metric values, indicating acceptable fusion performance for Res2NetFuse\_1. Thus, even when based on a single natural image, the Res2Net-based fusion model can achieve excellent fusion performance.

The comparisons in both reconstruction and fusion above validate the effectiveness of our training strategy. Now, let's summarize several reasons for its success:

1) Our method employs a lightweight network, distinct from deep networks in other computer vision tasks. 
As a result, it necessitates a smaller amount of training data. Several deep learning-based image fusion techniques predominantly leverage deep networks for the purpose of reconstruction, which entails a reduced training data requirement when compared to classification tasks.

2) Natural images inherently display substantial internal data repetition and robust predictive power \cite{shocher2018zero}, making them conducive to training a lightweight network with deep learning's strong generalization capabilities.

3) The significance of multi-scale feature representations in Res2Net cannot be overstated for our method. These representations facilitate the decoder's task of image reconstruction.

Considering these factors, we select Res2NetFuse\_1 as the final fusion model, which will be trained using a single natural image. Subsequent references to Res2NetFuse imply training with a single image.

\subsection{Assessment from Subjective and Objective Perspectives}
\label{Assessment}
We apply our method to 20 test images and present the visual results in Fig.~\ref{results2}.

Upon visual inspection, the fused results obtained by DCHWT exhibit noticeable noise and lack clarity in salient features. In contrast, the fused results from JSR and JSRSD appear to sharpen salient features excessively. Furthermore, FusionGAN and AT-GAN exhibit instability and introduce significant spectral contamination, as evidenced in  Fig.~\ref{results2}(g) and Fig.~\ref{results2}(i).

Conversely, Res2NetFuse, AEFusion, and DenseFuse produce results that closely align with human visual standards. MSDNet, on the other hand, tends to weaken salient features. CUFD and MUFusion manage to preserve texture details but introduce numerous artifacts into the fused results. Notably, Res2NetFuse delivers more stable results with enhanced salient features.

To objectively assess the proposed method's effectiveness, we employ various indices for evaluating the fused results, as displayed in Table \ref{tab:ob}.

In Table\ref{tab:ob}, the performance rankings are color-coded. Res2NetFuse exhibits advantages in these fusion evaluation metrics.

Notably, our method excels with the highest values in $Q_{abf}$, $FMI_w$, and SCD, signifying its capacity to retain more salient and visual information from the source images while reducing noise. Furthermore, it attains the second-highest scores in MS\_SSIM and $FMI_{dct}$ and the third-highest score in $SSIM_a$, highlighting its effectiveness in preserving structural information. Comparing Res2NetFuse with MSDNet illuminates Res2NetFuse's superior capability in extracting robust deep features. This enhancement significantly augments its aptitude for intricate image fusion tasks, establishing our method as a robust architecture for fusion.

\section{Conclusion}
\label{conclusion}
Leveraging the multi-scale Res2Net backbone, our architecture achieves breakthrough results in image fusion, demonstrating its efficacy in capturing nuanced features across diverse scales. Our findings indicate that training the network for image fusion can be effectively accomplished using just a single natural image. We also provided insights into the success of our proposed method. To enhance our fusion strategy, we introduced an attention model that generates more effective weight maps for fusing salient features from source images. Our approach has been demonstrated to outperform SOTA methods, as indicated by experimental results. In summary, this article primarily focuses on Res2Net's capability for multi-scale feature extraction in images. In future work, our research will pivot towards innovative global feature extraction in images.

Researchers contribute to these emerging prospects by conducting empirical studies, proposing novel algorithms, developing practical applications, and critically analyzing the implications and challenges associated with the future of image fusion. Each of these directions offers exciting opportunities for further exploration and advancement in the field.

\bibliographystyle{IEEEtran}
\bibliography{mybibfile.bib}

\end{document}